\documentclass[11pt,a4paper]{article}
\usepackage[hyperref]{naaclhlt2019}
\usepackage{times}
\usepackage{latexsym}
\usepackage{graphicx}
\usepackage{booktabs}
\usepackage{multirow}
\usepackage{subcaption}
\usepackage{amsmath}

\usepackage{url}

\aclfinalcopy 
\def\aclpaperid{2241} 

\title{Understanding language-elicited EEG data by predicting it from a fine-tuned language model}

\author{Dan Schwartz \\
  Carnegie Mellon University  \\
  5000 Forbes Ave \\
  Pittsburgh, PA 15213 \\
  {\tt drschwar@cs.cmu.edu} \\\And
  Tom Mitchell \\
  Carnegie Mellon University  \\
  5000 Forbes Ave \\
  Pittsburgh, PA 15213 \\
  {\tt tom.mitchell@cs.cmu.edu} \\}

\date{}

\begin{document}
\maketitle
\begin{abstract}
  Electroencephalography (EEG) recordings of brain activity taken while participants read or listen to language are widely used within the cognitive neuroscience and psycholinguistics communities as a tool to study language comprehension. Several time-locked stereotyped EEG responses to word-presentations -- known collectively as event-related potentials (ERPs) -- are thought to be markers for semantic or syntactic processes that take place during comprehension. However, the characterization of each individual ERP in terms of what features of a stream of language trigger the response remains controversial. Improving this characterization would make ERPs a more useful tool for studying language comprehension. We take a step towards better understanding the ERPs by fine-tuning a language model to predict them. This new approach to analysis shows for the first time that all of the ERPs are predictable from embeddings of a stream of language. Prior work has only found two of the ERPs to be predictable. In addition to this analysis, we examine which ERPs benefit from sharing parameters during joint training. We find that two pairs of ERPs previously identified in the literature as being related to each other benefit from joint training, while several other pairs of ERPs that benefit from joint training are suggestive of potential relationships. Extensions of this analysis that further examine what kinds of information in the model embeddings relate to each ERP have the potential to elucidate the processes involved in human language comprehension.
\end{abstract}

\section{Introduction}

\begin{figure}[t!]
  \centering
  \includegraphics[width=0.4\textwidth]{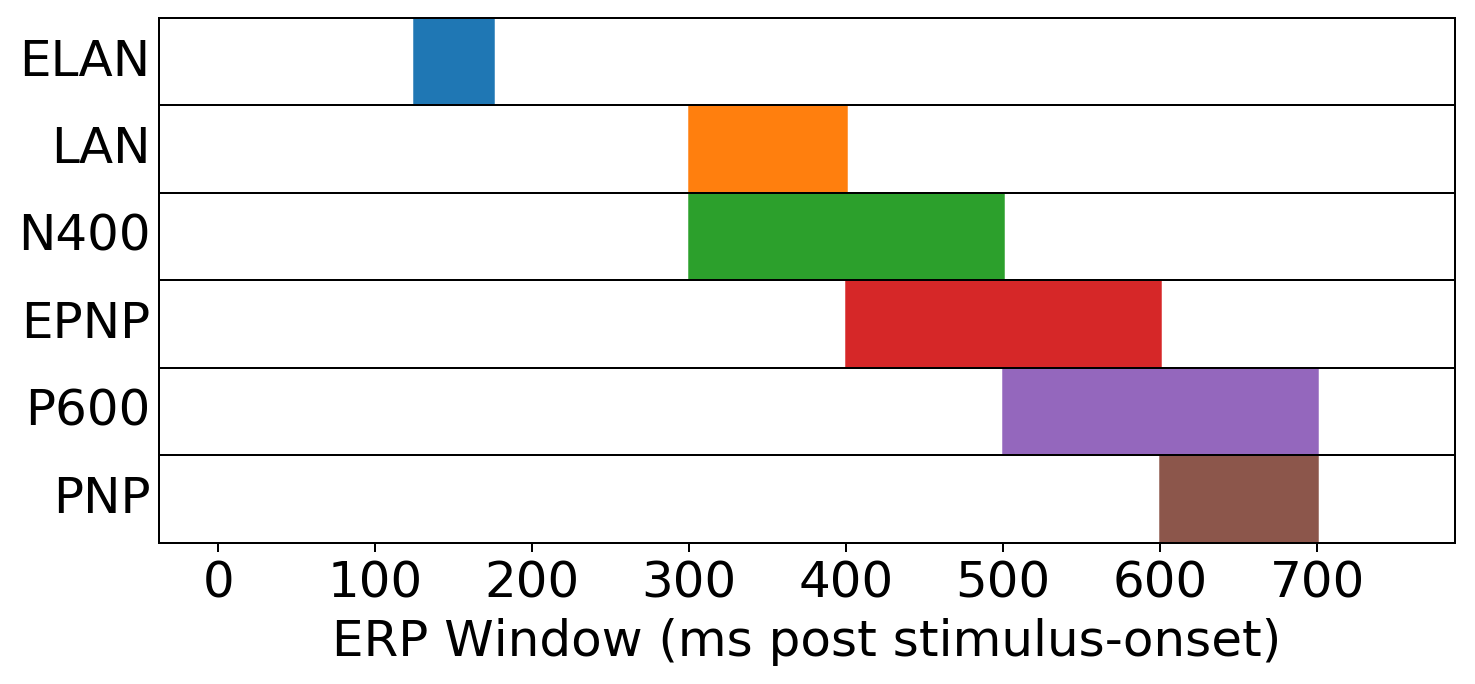}
  \includegraphics[width=0.4\textwidth]{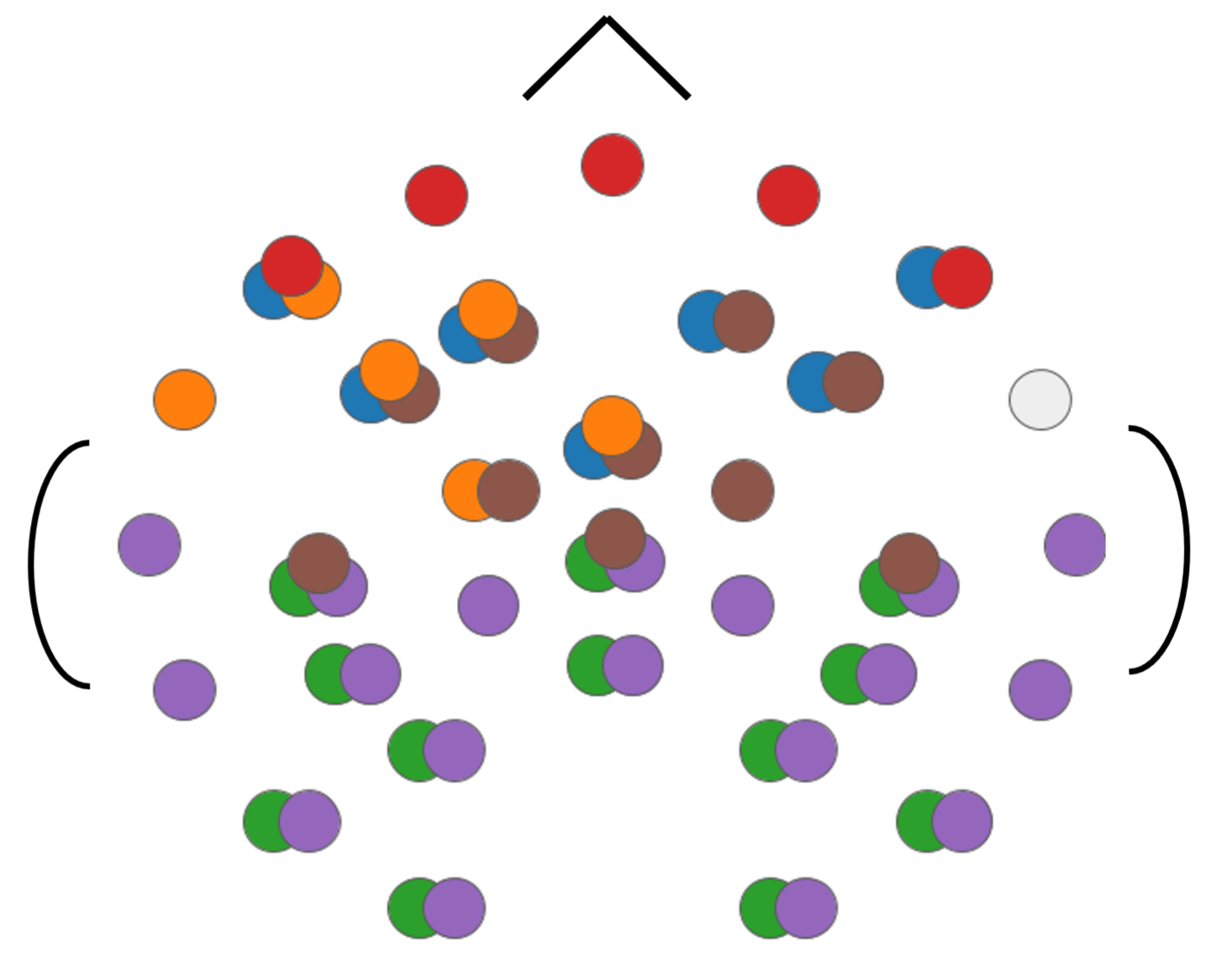}
  \caption[ERP locations and timing.]{The electrodes from which each event-related potential was recorded in the data from \citet{frank2015erp} (after figure 3 in \cite{frank2015erp}). The bottom portion of the figure shows a top-down schematic of the electrode locations with the nose facing towards the top of the page. Each ERP is the mean potential from all of the indicated electrodes during a specific time-window, creating a single scalar value per ERP per word. Overlapping circles indicate multiple ERPs recorded from the same electrode. The ELAN is measured from 125-175ms after stimulus onset, the LAN from 300-400ms, the N400 from 300ms-500ms, the EPNP from 400-600ms, the P600 from 500-700ms, and the PNP from 600-700ms.}
  \label{fig:erp}
\end{figure}

The cognitive processes involved in human language comprehension are complex and only partially identified. According to the dual-stream model of speech comprehension \cite{hickok2007cortical}, sound waves are first converted to phoneme-like features and further processed by a ventral stream that maps those features onto words and semantic structures, and a dorsal stream that (among other things) supports audio-short term memory.
The mapping of words onto meaning is thought to be subserved by widely distributed regions of the brain that specialize in particular modalities --- for example visual aspects of the word \textit{banana} reside in the occipital lobe of the brain and are activated when the word \textit{banana} is heard \cite{kemmerer2014cognitive} --- and the different representation modalities are thought to be integrated into a single coherent latent representation in the anterior temporal lobe \cite{ralph2010coherent}. While this part of meaning representation in human language comprehension is somewhat understood, much less is known about how the meanings of words are integrated together to form the meaning of sentences and discourses. One tool researchers use to study the integration of meaning across words is electroencephelography (EEG), which measures the electrical activity of large numbers of neurons acting in concert. EEG has the temporal resolution necessary to study the processes involved in meaning integration, and certain stereotyped electrical responses to word presentations, known as event-related potentials (ERPs), have been identified with some of the processes thought to contribute to comprehension.

\vspace{-4pt}

In this work, we consider six ERP components that have been associated in the cognitive neuroscience and psycholinguistics literature with language processing and which we analyze in the data from \citet{frank2015erp} (see Figure \ref{fig:erp} for spatial and temporal definitions of these ERP components). Three of these --- the N400, EPNP, and PNP responses --- are primarily considered markers for semantic processing, while the other three --- the P600, ELAN, and LAN responses --- are primarily considered markers for syntactic processing. However, the neat division of the ERP responses into either semantic or syntactic categories is controversial. The N400 response has been very well studied (for an overview see \cite{kutas2011thirty}) and it is well established that it is associated with semantic complexity, but the features of language that trigger the other ERP responses we consider here are poorly understood. We propose to use a neural network pretrained as a language model to probe what features of language drive these ERP responses, and in turn to probe what features of language mediate the cognitive processes that underlie human language comprehension, and especially the integration of meaning across words.

\section{Background}

While a full discussion of each ERP component and the features of language thought to trigger each are beyond the scope of this document (for reviews see e.g. \citet{frank2015erp}, \citet{kemmerer2014cognitive}, \citet{kutas2011thirty}, \citet{kuperberg2003electrophysiological}, and \citet{van2012prediction}), we introduce some basic features of ERP components to help in the discussion later. ERP components are electrical potential responses measured with respect to a baseline that are triggered by an event (in our case the presentation of a new word to a participant in an experiment). The name of each ERP component reflects whether the potential is positive or negative relative to the baseline. The N400 is so-named because it is \textbf{N}egative relative to a baseline (the baseline is typically recorded just before a word is presented at an electrode that is not affected by the ERP response) and because it peaks in magnitude at about \textbf{400}ms after a word is presented to a participant in an experiment. The P600 is \textbf{P}ositive relative to a baseline and peaks around \textbf{600}ms after a word is presented to a participant (though its overall duration is much longer and less specific in time than the N400). The post-N400 positivity is so-named because it is part of a biphasic response; it is a positivity that occurs after the negativity associated with the N400. The early post-N400 positivity (EPNP) is also part of a biphasic response, but the positivity has an eariler onset than the standard PNP. Finally, the LAN and ELAN are the left-anterior negativity and early left-anterior negativity respectively. These are named for their timing, spatial distribution on the scalp, and direction of difference from the baseline. It is important to note that ERP components can potentially cancel and mask each other, and that it is difficult to precisely localize the neural activity that causes the changes in electrical potential at the electrodes where those changes are measured.

\section{Related Work}

This work is most closely related to the paper from which we get the ERP data: \citet{frank2015erp}. In that work, the authors relate the surprisal of a word, i.e. the (negative log) probability of the word appearing in its context, to each of the ERP signals we consider here. The authors do not directly train a model to predict ERPs. Instead, models of the probability distribution of each word in context are used to compute a surprisal for each word, which is input into a mixed effects regression along with word frequency, word length, word position in the sentence, and sentence position in the experiment. The effect of the surprisal is assessed using a likelihood-ratio test. In \citet{hale2018finding}, the authors take an approach similar to \citet{frank2015erp}. The authors compare the explanatory power of surprisal (as computed by an LSTM or a Recurrent Neural Network Grammar (RNNG) language model) to a measure of syntactic complexity they call ``distance" that counts the number of parser actions in the RNNG language model. The authors find that surprisal (as predicted by the RNNG) and distance are both significant factors in a mixed effects regression which predicts the P600, while the surprisal as computed by an LSTM is not. Unlike \citet{frank2015erp} and \citet{hale2018finding}, we do not use a linking function (e.g. surprisal) to relate a language model to ERPs. We thus lose the interpretability provided by the linking function, but we are able to predict a significant proportion of the variance for all of the ERP components, where prior work could not. We interpret our results through characterization of the ERPs in terms of how they relate to each other and to eye-tracking data rather than through a linking function.
The authors in \citet{wehbe-EtAl:2014:EMNLP2014} also use a recurrent neural network to predict neural activity directly. In that work the authors predict magnetoencephalography (MEG) activity, a close cousin to EEG, recorded while participants read a chapter of \textit{Harry Potter and
the Sorcerer’s Stone} \cite{rowling1999harry}. Their approach to characterization of processing at each MEG sensor location is to determine whether it is best predicted by the context vector of the recurrent network (prior to the current word being processed), the embedding of the current word, or the probability of the current word given the context. In future work we also intend to add these types of studies to the ERP predictions.

\section{Method}

\paragraph{Data.} We use two sources of data for this analysis. The primary dataset we use is the ERP data collected and computed by \citet{frank2015erp}, and we also use behavioral data (eye-tracking data and self-paced reading times) from \citet{frank2013reading} which were collected on the same set of 205 sentences. In brief, the sentences were selected from sources using British English with a criterion that they be understandable out of context. We use the ERP component values as computed by \citet{frank2015erp} which have been high-pass filtered at 0.5 Hz to reduce correlation between ERP components and modulus transformed \cite{john1980alternative} to make the distribution of component values more normal. We do not use the 100ms pre-trial baseline which is made available by \citet{frank2015erp} and which they use as a separate input to the mixed effects regression. For more information about the ERP datasets and data collection procedures we refer the reader to the original papers. For the behavioral data, we use self-paced reading times and four eye-tracking measures. Self-paced reading time is considered a signal of integration difficulty (i.e. as it becomes more difficult to integrate the meaning of the current word into the context, the amount of time a reader spends on the current word increases). The eye-tracking measures are intended to capture both early effects (effects modulated primarily by properties of the word independent of its context, such as word frequency and word length) and late effects (effects modulated by the context in which the word is found, i.e. comprehension difficulty) in word processing \cite{rayner2006eye_control}. In both cases, the eye-tracking measures provide a signal of overt visual attention, which is thought to strongly correlate with covert perceptual attention in normal reading \cite{rayner2009eye}. We log-transform the self-paced reading time and the eye-tracking measures.

\begin{figure}[t!]
  \centering
  \includegraphics[width=0.49\textwidth]{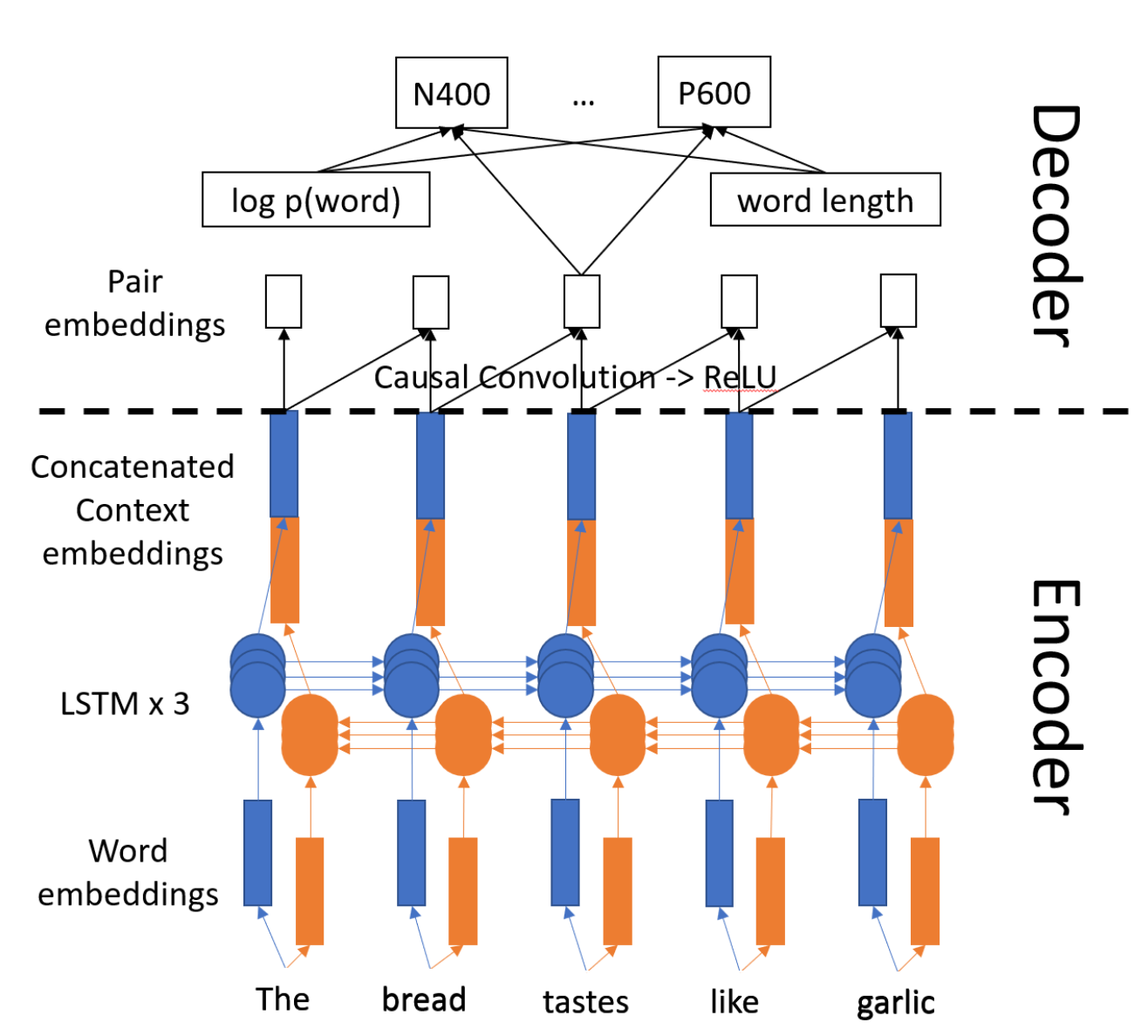}
  \caption[Model architecture.]{The model uses an encoder based on the architecture and regularization in \citet{merity2017regularizing} and pretrained by \citet{howard2018fine}. Within this architecture 2 independent 3-layer LSTM models encode a sentence. The context-embeddings output from each encoder are then concatenated together to give a single representation to each word in the sentence. These concatenated context-embeddings are fed into a causal-convolution, which learns a function to combine each pair of context-representations into a pair-embedding. A rectified linear unit (ReLU) non-linearity is applied to the pair-embedding, after which independent linear layers map the pair-embedding along with the log-probability of a word and the word-length to a prediction of each ERP or behavioral signal.}
  \label{fig:architecture}
\end{figure}

\paragraph{Model.} To predict the ERP signals in the data, we start with a 3-layer bidirectional LSTM-based language model encoder using the architecture found in \citet{merity2017regularizing} and pretrained on the WikiText-103 dataset \cite{merity2016pointer} (we use the pretrained model from \citet{howard2018fine}). The pretraining objective is to minimize the negative log-likelihood of the next word for the forward LSTM and the previous word for the reverse LSTM. The word-embeddings (input embeddings) in the encoder have 400 components, the hidden layer outputs have 1150 components each, and the context-embeddings output from the encoder have 400 components. The forward-encoder and backward-encoder are independently fine-tuned on the baby version of the British National Corpus \cite{bncbaby} to help with prediction of British English (both the ERP data and eye-tracking data use British English). During task training the two encoders' output embeddings are concatenated together and fed into a causal-convolution layer which combines each pair of adjacent timepoints into a single pair-embedding with 10 components. The causal-convolution (i.e. convolution which is left padded) ensures that the pair-embeddings are aligned so that the prediction targets correspond to the later word in the pair. In other words the pair can be thought of as representing the `current' and `previous' words together. A ReLU is applied to the pair-embedding before it, along with the word length and the log probability of the word, is fed into a linear output layer to predict each ERP and behavioral measure (see Figure \ref{fig:architecture}). The convolution and linear layers are initialized using the default PyTorch \cite{paszke2017automatic} initialization, i.e. the initialization proposed in \citet{he2015delving}. The encoder portion of the model includes dropout as applied in \citet{merity2017regularizing}, but we use different dropout probabilities when we fit the neural and behavioral data (the dropout probability on the input embeddings was 0.05, 0.4 on the input to the LSTM, 0.4 on LSTM hidden layers, 0.5 on the output of the LSTM, and 0.5 on the recurrent weights). We did not find dropout in the decoder to be helpful. We use the Adam optimizer \cite{kingma2014adam} with $\beta_1 = 0.95, \beta_2 = 0.999$ for training and we use mean squared error as the loss.

\paragraph{Procedure.} We begin our training procedure by fine-tuning the forward- and backward-encoders independently on the baby version of the British National Corpus \cite{bncbaby}. This corpus has British English that may help in modeling the University College London corpus, while not overlapping with it.

After the model fine-tuning, we estimate how well the model predicts each of the ERP signals and eye-tracking measures by training the model 100 times with different train/test splits and decoder parameter initializations. We use 10\% of the data for testing and the remainder for training. The sentences in the ERP data are split at random. After we split the data, we compute the mean and standard deviation of each ERP signal (and each eye-tracking measure and the self-paced reading time) within participant on the training data. We use these values to standardize the training data within participant, and then average the data from all of the participants together. After we average, we again compute the mean and standard deviation to standardize the average. We follow a similar procedure for the test data, but we use the mean and standard deviation from the training data when standardizing. Note that we use the log of the behavior measures, and the log is taken before the data-standardization.

In the loss function (and when we evaluate model performance) we only consider content words. We mark as a content word any word that is an adjective, adverb, auxiliary verb, noun, pronoun, proper noun, or verb (including to-be verbs). All other words are considered function words.

During the first 20 epochs of training, only the parameters of the decoder are modified. Following this, we train the model for an additional 15 epochs during which the parameters of the decoder and the final layer of the encoder (the final LSTM layer in both the forward and backward encoder) can be modified. We also experimented with additional training epochs and allowing all parameters of the model to be modified, but we found that this caused overfitting.

\paragraph{Comparing models trained with different loss functions.} To better understand the relationship between ERP signals, and between ERP signals and behavioral data, we train the model with different loss functions that include mean squared error terms corresponding to various combinations of the ERP signals and behavioral data. For example, one of the training variations includes a mean squared error term for the P600 and a mean squared error term for the N400 in the loss, but does not use the other signals during training. In this variation, for a mini-batch of size $B$, where example $b$ has $T_b$ content tokens and the superscripts $p$ and $a$ denote the predicted and actual values for a measure respectively, the loss function can be written as:
\begingroup
\setlength\abovedisplayskip{8pt}
\setlength\belowdisplayskip{8pt}
\begin{equation}
\begin{split}
    &\frac{1}{\sum_{b=1}^B T_b} \sum_{b=1}^B \sum_{t=1}^{T_b} (\mathrm{P600}^{p}_{b,t} - \mathrm{P600}^{a}_{b,t})^2 \\
    &\qquad\qquad\qquad\quad + (\mathrm{N400}^{p}_{b,t} - \mathrm{N400}^{a}_{b,t})^2
\end{split}
\end{equation}
\endgroup
For each of the training variations, we repeat the training procedure described above (but fine-tuning the language model on the British National Corpus is done only once). We use a consistent train/test split procedure, such that the split for the $ith$ run of the 100 runs is the same across all training variations, but the split changes between run $i$ and run $j$. This enables us to use paired statistical testing when we test for significance.

We test for whether the proportion of variance explained (computed as $1 - \frac{\text{MSE}}{\text{variance}}$ on the validation set) on each ERP and behavioral measure is significantly different from 0 using the single sample t-test controlled for false discovery rate using the Benjamini-Hochberg-Yekutieli procedure \cite{benjamini2001control} with a false discovery rate of 0.01.

To test whether the proportion of variance explained is different between different training variations (for example training with just the N400 signal included in the loss vs. training with both the N400 and the LAN included in the loss), we use a paired t-test. We then adjust for the false discovery rate again with a rate of 0.01.

\section{Results}

\paragraph{All ERP components are predictable.} In the original study on this dataset, the investigators found that when surprisal was used as a linking function between the language model and the mixed effects regression, the only ERP for which the surprisal showed a significant effect in the regression was the N400 \cite{frank2015erp}. In contrast, we find that when we directly predict the ERP signals we are able to predict a significant proportion of the variance for all of them (see Table \ref{table:erp_table}).

\begin{table*}
\centering
\begin{tabular}{l l c | l l c | l l c}
\toprule
Target & Additional & POVE & Target & Additional & POVE & Target & Additional & POVE \\
\midrule
ELAN & & 0.20 & LAN & & 0.30 & N400 & & 0.26\\
ELAN & + EPNP & 0.22 & LAN & + EPNP & 0.31 & & & \\
ELAN & + N400 & 0.22 & LAN & + PNP & 0.32 & & & \\
ELAN & + PNP & 0.22 & LAN & + P600 & 0.32 & & & \\
ELAN & + P600 & 0.22 & LAN & + PNP, N400 & 0.33 & & &\\
\midrule
EPNP & & 0.34 & P600 & & 0.27 & PNP & & 0.33 \\
EPNP & + LAN & 0.35 & P600 & + EPNP & 0.30 & PNP & + LAN & 0.36 \\
EPNP & + GROUP A & 0.36 & P600 & + LAN & 0.30 & PNP & + GROUP B & 0.36 \\
\bottomrule
\end{tabular}
\caption{Proportion of variance explained (POVE) for each of the ERP components (mean of 100 training runs). The second column in each cell shows which ERP components in addition to the target ERP component were included in training. All combinations of training signals were explored. Shown is the best combination for each ERP target as well as
every combination which is (i) significantly different from training on the target component alone, (ii) not significantly different from the best training combination, and (iii) uses no more than the number of signals used by the best combination. The N400 is predicted best when only the N400 signal is included in training. All values are significantly different from 0. GROUP A refers to (PNP, ELAN, LAN, P600) and GROUP B refers to (EPNP, ELAN, LAN, P600).}
\label{table:erp_table}
\end{table*}

\paragraph{Joint training benefits ERP component prediction.} To explore the relationship between ERP components, we train $63 = {6\choose1} + {6\choose2} + \cdots + {6\choose6}$ different models using all of the possible combinations of which of the six ERP signals are included in the loss function during training. For each of the six ERP components, we look for the best performing models (see Table \ref{table:erp_table}). The N400 is best predicted when the model is trained on that component independently, but every other ERP component prediction can be improved by including a second ERP component in the training. Thus multitask learning has a clear benefit when applied to the ERP data and some information is shared between ERP component predictions via the model parameters. We also note that it is not the case that training with more ERP components is always better, or that the signals which are most correlated benefit each other most (see Appendix \ref{sec:appendix}). The relationship between components clearly impacts whether the prediction of one ERP component benefits from the inclusion of others in model training. The results suggest that 8 pairs of ERP signals are related to each other: the LAN is paired with the P600, EPNP, and PNP, the ELAN with the N400, EPNP, PNP, and P600, and the EPNP is paired with the P600. We discuss these relationships in the Discussion section.

\begin{figure*}
  \centering
  \begin{subfigure}[b]{0.3\textwidth}
    \includegraphics[width=\textwidth]{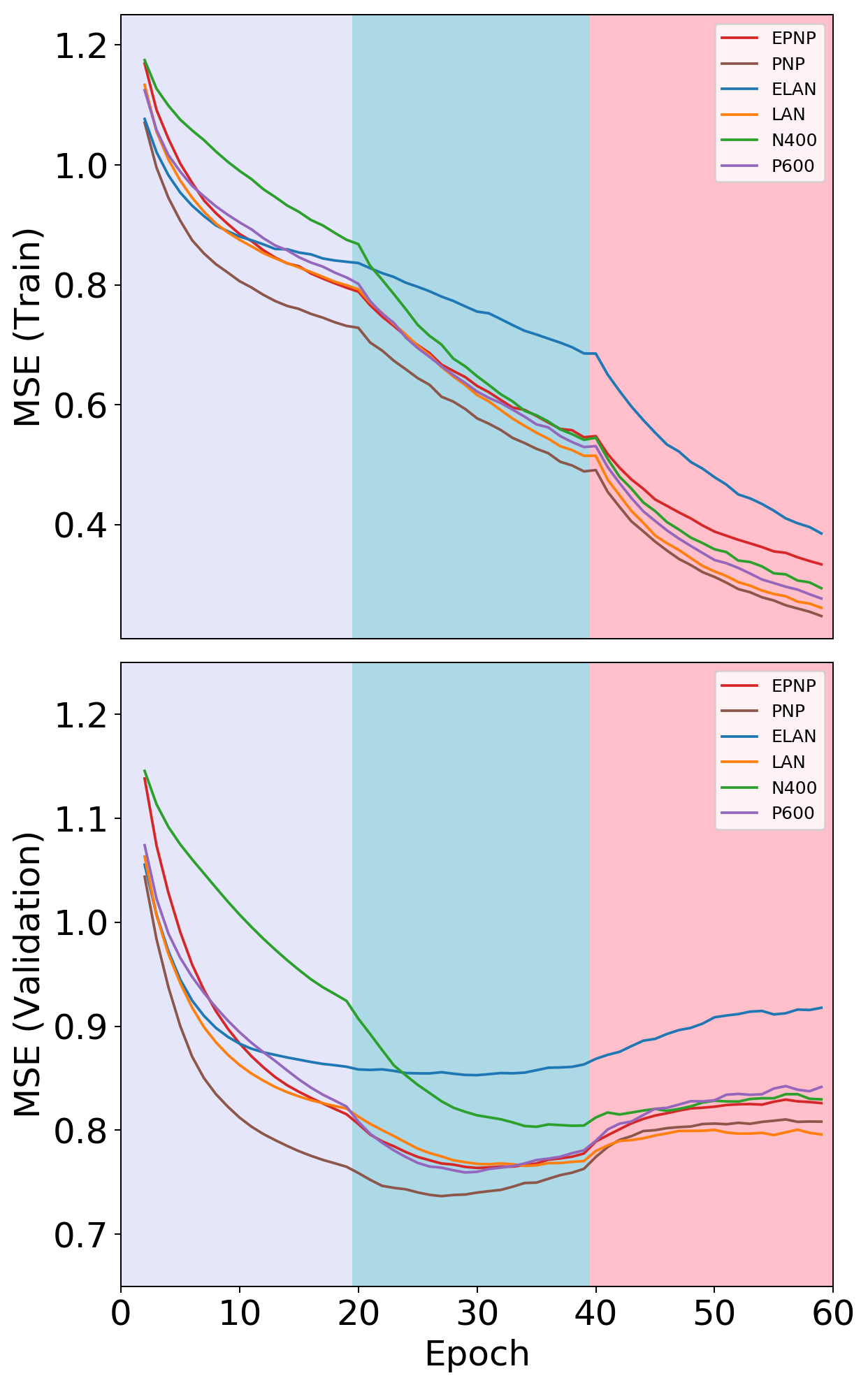}
    \label{fig:train_curves_single}
    \caption{Independently trained}
  \end{subfigure}             
  \begin{subfigure}[b]{0.3\textwidth}
    \includegraphics[width=\textwidth]{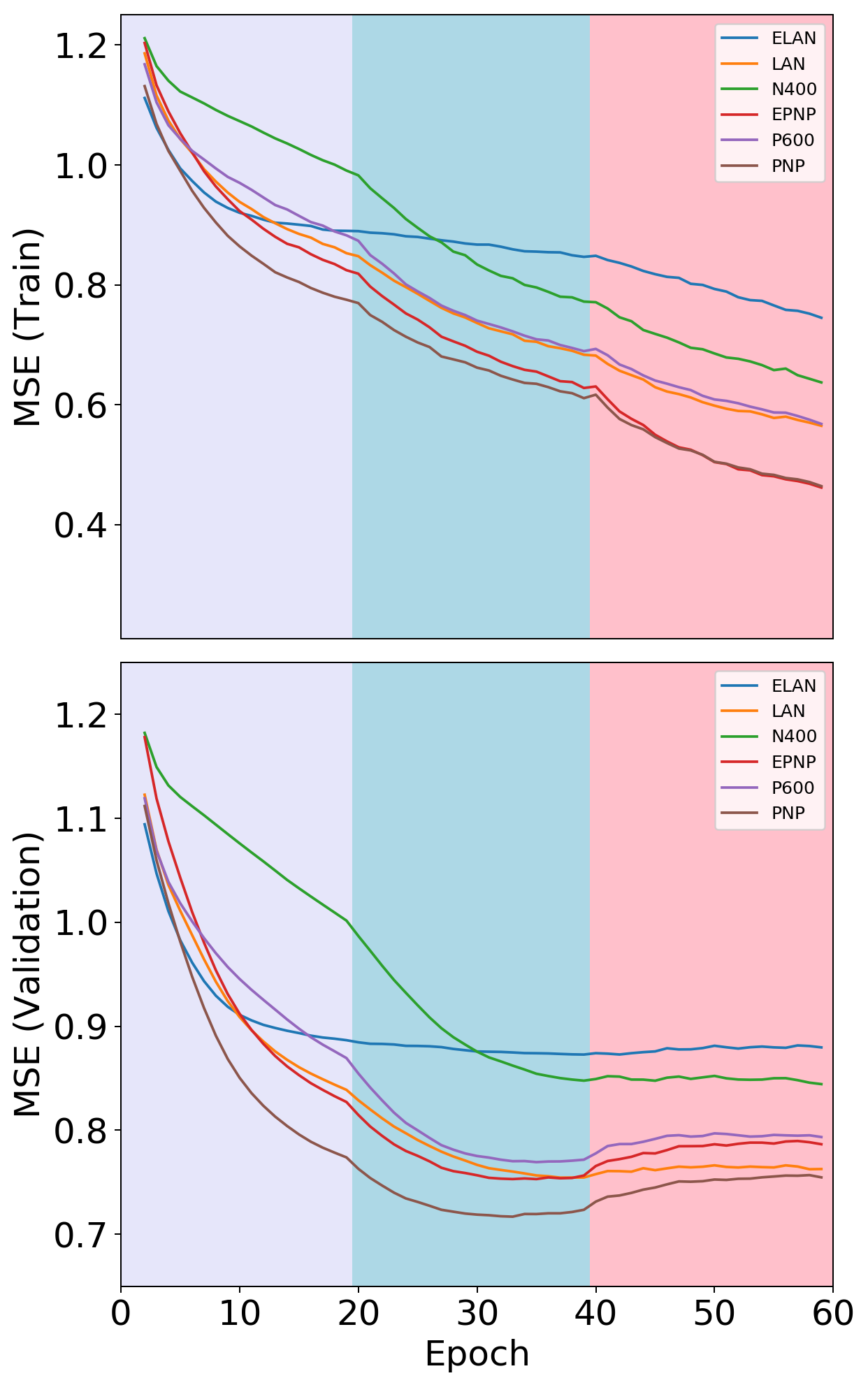}
    \label{fig:train_curves_all}
    \caption{Jointly trained}
  \end{subfigure}             
  \begin{subfigure}[b]{0.3\textwidth}
    \includegraphics[width=\textwidth]{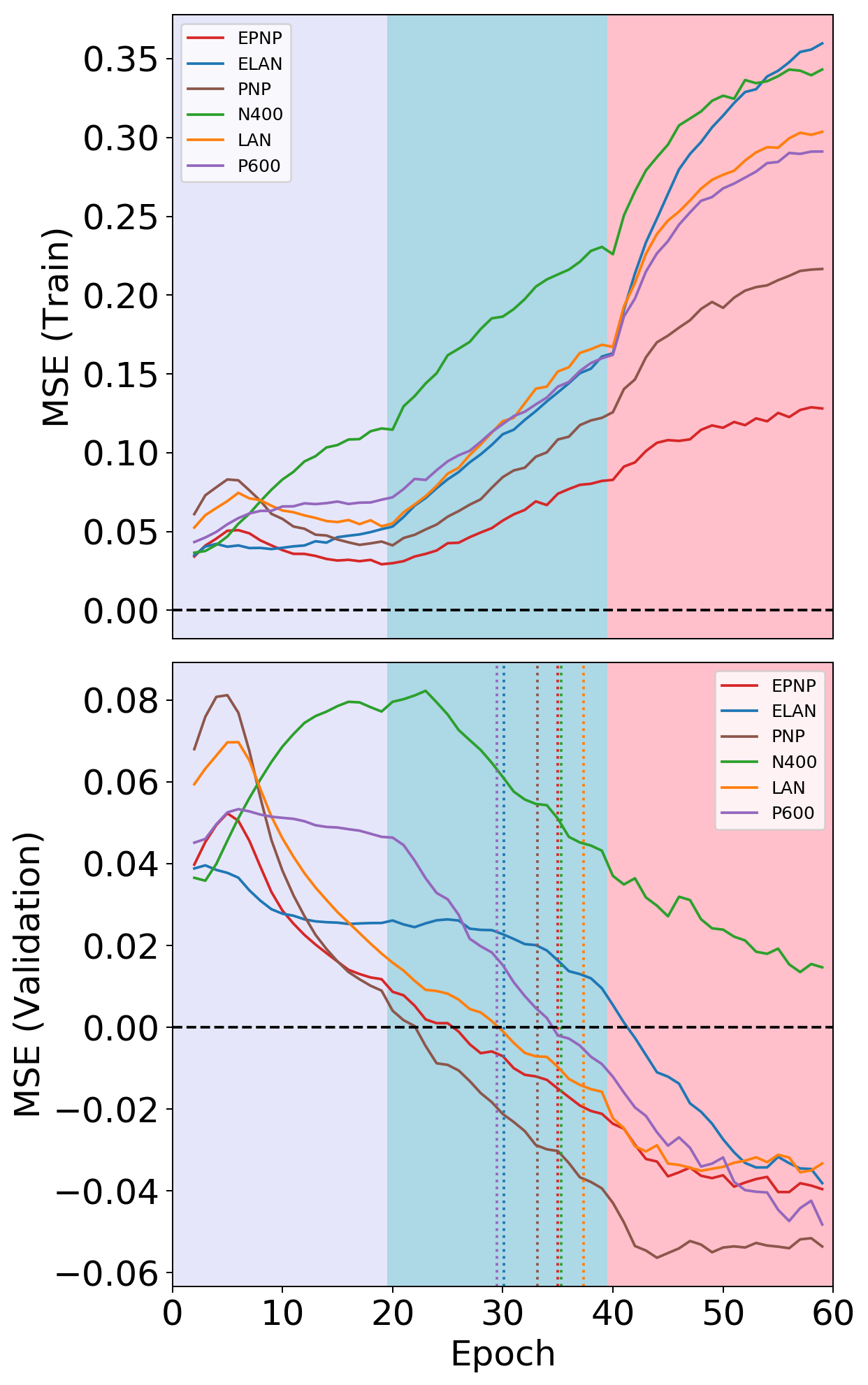}
    \label{fig:train_curves_diff}
    \caption{Joint - Indep.}
  \end{subfigure}
  \caption[ERP train/test curves.]{The mean squared error (MSE) for prediction of each of the ERP signals during each epoch of training (mean of 100 training runs). The first 2 epochs have been omitted for clarity. During the first 20 epochs (lavender background), only the decoder parameters are modified. During the next 20 epochs (light blue background), the parameters in the final layer of the encoder are also modified. During the last 20 epochs (pink background), all of the parameters are modified. Note that in this model architecture, information can be shared between ERP signals even when only the decoder is modified. The figure shows the MSE when separate models are trained for each ERP independently (a), the MSE when a single model is trained on all ERPs jointly (b), and the difference between these two scenarios (c). The top row in each column shows the MSE on the training data while the bottom row shows the MSE on the validation data. In the bottom row right, the dotted vertical lines indicate the epoch at which the minimum MSE is reached in the lower of the independent or joint training. The LAN, EPNP, and PNP all show modest benefits from joint training before overfitting sets in (the minimum value occurs in the joint training scenario), while all ERP signals other than the N400 show reduced overfitting in joint training.}
  \label{fig:erp_train_curves}
\end{figure*}

In an additional analysis, we modified our training procedure slightly to probe how jointly training on multiple ERP components compares to training individually on each ERP component. In this analysis we compare only training on each ERP component individually to training on all six ERP components together. We also train for a total of 60 epochs (rather than the 35 epochs used elsewhere). During the first 20 epochs we allow only the parameters of the decoder to be modified. During the next 20 epochs, we allow the parameters of the decoder and the final layer of the encoder (i.e. the final recurrent layer) to be modified. During the last 20 epochs, we allow all of the parameters of the model to be modified. The mean squared error for each of the ERP components from this analysis is shown for each epoch in Figure \ref{fig:erp_train_curves}. From the loss curves, we make a few observations. First, we see inflection points at epochs 20 and 40, when we allow more parameters of the model to be modified. The first inflection point indicates that allowing the recurrent layer to be modified benefits the prediction, while the second inflection point shows that overfitting becomes more severe if we allow all parameters of the model to be modified. We also see from these curves that part of the benefit of joint training is that it helps reduce overfitting -- we see less of a climb in the validation loss after the minimum point in the joint training. Beyond this reduction in overfitting severity, we note that for some of the ERP components (the LAN, EPNP and PNP components) joint training actually gives a better overall minimum in prediction error.

\paragraph{Behavioral data benefits the prediction of ERP components.} We are also interested in whether behavioral data can be used to improve ERP prediction since it should signal both the amount of overt attention required at various points in a sentence as well as integration difficulty. To study this question, we again train models using different combinations of training signals that include or do not include the behavioral data predictions in the loss function (see Table \ref{table:behav_table}). We see that self-paced reading time indeed can improve prediction of a target ERP component relative to training on the target ERP component alone by about the same amount as the best combination of ERP components for all but the N400. Eye-tracking data can also improve the prediction accuracy of the ELAN, P600, and PNP components.

\begin{table*}
\centering
\begin{tabular}{l l c | l l c | l l c}
\toprule
Target & Additional & POVE & Target & Additional & POVE & Target & Additional & POVE \\
\midrule
ELAN & & 0.20 & LAN & & 0.30 & N400 & & 0.26 \\
ELAN & + ERP & \textbf{0.22} & LAN & + ERP & \textbf{0.33} & N400 & + ERP & 0.26 \\
ELAN & + READ & \textbf{0.22} & LAN & + READ & \textbf{0.31} & N400 & + READ & 0.27 \\
ELAN & + EYE & \textbf{0.22} & LAN & + EYE & 0.30 & N400 & + EYE & 0.25 \\
\midrule
EPNP & & 0.34 & P600 & & 0.27 & PNP & & 0.33 \\
EPNP & + ERP & \textbf{0.36} & P600 & + ERP & \textbf{0.30} & PNP & + ERP & \textbf{0.36} \\
EPNP & + READ & \textbf{0.35} & P600 & + READ & \textbf{0.29} & PNP & + READ & \textbf{0.34} \\
EPNP & + EYE & 0.34 & P600 & + EYE & \textbf{0.29} & PNP & + EYE & \textbf{0.34} \\
\bottomrule
\end{tabular}
\caption{Proportion of variance explained (POVE) for each of the ERP components (mean of 100 training runs). +ERP indicates the best combination of ERP training signals for the target ERP component, + READ indicates the inclusion of self-paced reading times, +EYE indicates the inclusion of eye-tracking data, and bold font indicates a significant difference from training on the target component alone.}
\label{table:behav_table}
\end{table*}

\paragraph{Insensitivity to choice of architecture.} One potential concern about our results is the degree to which the relationships we see between ERP components and between ERP components and behavioral data is an artefact of our rather arbitrary choice of network architecture. We partially address this by running the same analysis using (i) only the forward direction of the encoder, and (ii) only the word-embeddings (the input embeddings) and not the context-embeddings (the output embeddings) of the encoder. The proportion of variance explained for each ERP component is lower using these variants of the analysis than using the bidirectional variant (see Appendix \ref{sec:appendix}), but qualitatively the relationships are similar. We leave further analysis of the sensitivity of our qualitative results to choice of architecture for future work.

\section{Discussion}

In this work we find that all six of the ERP components from \citet{frank2015erp} can be predicted above chance by a model which has been pretrained using a language modeling objective and then directly trained to predict the components. This is in contrast to prior work which has successfully linked language models to the N400 \cite{frank2015erp} and P600 \cite{hale2018finding} but not the other ERP components. We also note that contrary to \citet{hale2018finding}, we find that an LSTM does contain information that can be used to predict EEG data, and in particular that it can predict the P600. We speculate that the analysis used in \citet{hale2018finding} did not find reliable effects because the language models were related to the EEG data through functions chosen a priori (the surprisal, and the `distance' metric). These functions, though interpretable, might be interpretable at the cost of losing much of the information in the representations learned by the network. 

In addition, we show through our multitask learning analysis that information is shared between ERP components, and between ERP components and behavioral data. Although these relationships must be viewed with caution until they can be verified across multiple datasets and with more variation in neural network architectures, here we consider some potential reasons for our findings. The broad point we wish to make is that by better understanding which ERP components share information with each other and with behavioral data through the type of analysis we present here (multitask learning) or other means, we can better understand what drives each ERP component and in turn the processes involved in human language comprehension.

\paragraph{Relationship between ERPs.} Our findings that the LAN and P600 are related, and that the ELAN and P600 are related are expected from both a theoretical perspective and from previous work examining the interactions of ERP components \cite{gunter1997syntax,hagoort2003real,hahne1999electrophysiological,kutas2006psycholinguistics,palolahti2005event}. Since the ELAN and LAN have been theorized by some to mark word-category (i.e. part-of-speech) or morpho-syntactic (e.g. subject-verb number agreement) violations \cite{friederici2011brain,hahne2002differential,hagoort2003syntax} and the P600 is considered a marker for syntactic effort \cite{coulson1998expect,huettig2015four,kemmerer2014cognitive,kuperberg2007neural,kuperberg2003electrophysiological,van2012prediction}, these signals would naturally be related to each other.

The other relationships we find are more surprising. Some researchers have speculated that the LAN and ELAN are markers for working memory demands \cite{king1995did,kutas2006psycholinguistics}, and that indeed these might be part of sustained negativities that are frequently masked by the P600 \cite{kemmerer2014cognitive}. If we take this view, then we would expect to find them in the presence of semantic and syntactic complexity, and this might explain why they seem to benefit from joint training with the other ERP component signals (and benefit prediction of other ERP signals with which they are trained). However, it is notable that predictions of the LAN and ELAN do not benefit each other in our analysis, and that the N400 (a marker for semantic complexity) is not benefited by the prediction of any other ERP component. This absence is by no means definitive, but it undermines the argument that all of these relationships can be explained by complexity and working memory demands alone. 

The relative isolation of the N400 from other ERP components in our analysis is interesting. If the N400 is a marker for semantic memory retrieval \cite{kutas2011thirty}, then it might be expected to be somewhat isolated from the other components, which may involve syntactic processing or later integration effects.

Alternatively, the relationships we find in our analysis might be an artefact of the way the ERPs are operationalized in \citet{frank2015erp}. Several of the pairings we find overlap spatially and are near to each other in time, so the ERP components might spill over into each other. Further work is required to disambiguate between these possibilities.

\paragraph{Relationship between behavioral data and ERPs.} It is reassuring to see that jointly training models to predict behavioral data along with a target ERP component benefits the prediction of the ERP component compared to training on the target ERP component alone. The benefit to prediction in this case cannot be explained as an artefact of how the ERP components are operationalized in the datasetes we use for analysis. 

Self-paced reading times widely benefit ERP prediction, while eye-tracking data seems to have more limited benefit to just the ELAN, LAN, and PNP ERP components. It's difficult to know why this might be the case, but perhaps it is not a coincidence that these three ERP components also show up frequently in the pairs of components that benefit from joint training. If indeed the PNP marks semantic role irregularities \cite{van2012prediction} and the ELAN and LAN mark working memory or look-forward or look-back operations \cite{kutas2006psycholinguistics}, then its possible that eye-movements might be more related to these types of operations than to general semantic and syntactic complexities marked by other ERP components. Self-paced reading might better capture these generic difficulties. This explanation is highly speculative, and further work is required to determine whether the relationships between the ERP components and behavioral data are consistent across datasets, and if so, what the explanation is for these relationships.

\paragraph{Choice of bidirectional architecture.} We emphasize that the neural network architecture we chose for these analyses was motivated primarily by its success on downstream NLP tasks, public availability of pre-trained models and code, and prior work studying how best to fine-tune the model \cite{howard2018fine,merity2017regularizing}. We do not claim that this architecture reflects human processing. We experimented with a forward-only model variant of our analysis, and found that the bidirectional model predicts brain activity better than the forward-only version (see Appendix \ref{sec:appendix}). Although the bidirectional model has access to `future' language input, it does not have access to future brain-activity, so the bidirectional model is not `cheating' when it makes predictions. There are at least three possible explanations for why the bidirectional model performs better than the forward-only model. First, it is possible that when a human reads a sentence, he or she predicts the upcoming language input. Under this hypothesis, a model with access to the future language input can do a better job of predicting the current brain activity because the future language is reflected in that brain activity. Second, it is possible that a bidirectional model is simply able to produce better embeddings for each word in the input because it has more context than a forward-only model. For example, the bidirectional model might be (implicitly) better at anaphora resolution given more context. Under this hypothesis, the additional context given to the model partially compensates for its relative deficit of real-world knowledge compared to a human. Where a human can in many cases solve the anaphora resolution problem by using background knowledge and does not need to see the future language input, a model benefits from additional context. Finally, in our setup, the bidirectional model has more parameters than the forward-only model, and the additional degrees of freedom might give the model an advantage in predicting brain activity. Exploration of why the bidirectional model is better than the forward-only model is an interesting question, but it is left to future work. Additionally, as we noted earlier, the qualitative results of our analysis (e.g. how ERP components relate to each other) should be viewed with caution until they are replicated across multiple choices of architecture.

\section{Conclusion}

We have shown that ERP components can be predicted from neural networks pretrained as language models and fine-tuned to directly predict those components. To the best of our knowledge, prior work has not successfully used statistical models to predict all of these components. Furthermore, we have shown that multitask learning benefits the prediction of ERP components and can suggest how components relate to each other. At present, these joint-training benefit relationships are only suggestive, but if these relationships ultimately lead to insights about what drives each ERP component, then the components become more useful tools for studying human language comprehension. By using multitask learning as a method of characterization, we have found some expected relationships (LAN+P600 and ELAN+P600) and several more surprising relationships. We believe that this is exactly the kind of finding that makes multitask learning an interesting exploratory technique in this area. Additionally, we have shown that information can be shared between heterogeneous types of data (eye-tracking, self-paced reading, and ERP components) in the domain of human language processing prediction, and in particular between behavioral and neural data. Given the small datasets associated with human language processing, using heterogeneous data is a potentially major advantage of a multitask approach. In future work, we will further explore what information is encoded into the model representations when neural and behavioral data are used to train neural networks, and how these representations differ from the representations in a model trained on language alone.

\section{Acknowledgments}

We thank our reviewers for their valuable feedback. This work is supported in part by National Institutes of Health grant number U01NS098969.

\bibliography{naaclhlt2019}
\bibliographystyle{acl_natbib}

\clearpage

\appendix

\section{Appendix}
\label{sec:appendix}

\begin{figure*}
  \centering
  \includegraphics[width=0.9\textwidth]{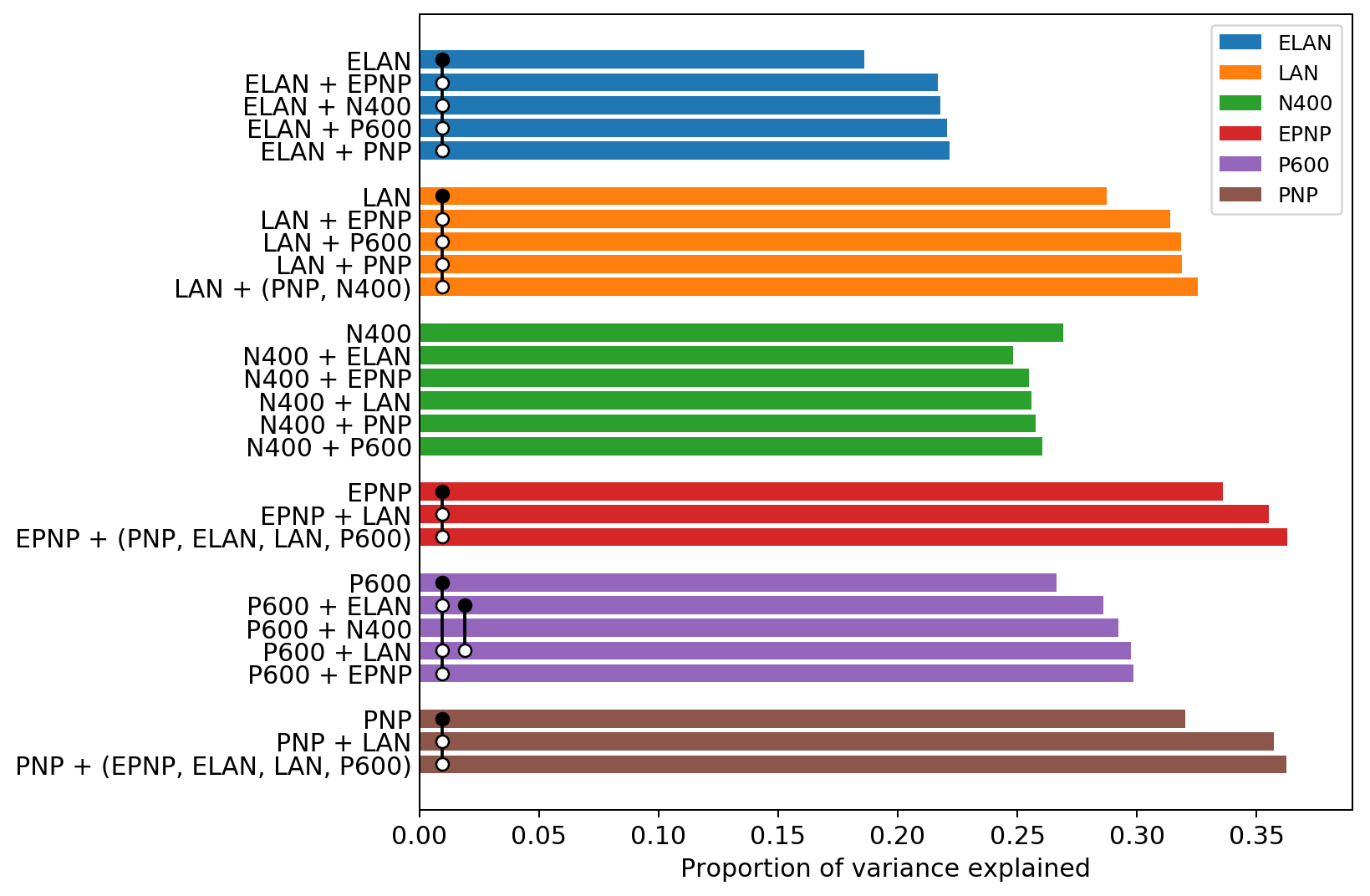}
  \caption[ERP training combinations.]{The proportion of variance explained for prediction of each of the ERP signals (mean of 100 training runs). The target ERP is indicated by color; each group of bars shows performance for a different target ERP. The top bar in each group shows the proportion of variance explained when the model is trained using only the target ERP. The bottom bar in each group shows the maximum proportion of variance explained over all combinations of training ERPs (or in the case of the N400, the second best). Also shown in each group are any training combinations that (i) used no more than the number of ERP signals used by the combination that achieved the maximum, and (ii) which were not significantly different from the maximum. Bars are statistically different from each other if a black dot on one bar is connected by a contiguous vertical line to a white dot on the other bar. The bars in the N400 group are not significantly different from each other. The N400 signal is best predicted when the model is trained on just that signal. In every other group, there is at least one ERP that, when combined with the target ERP during training, improves the prediction of the target ERP. The results suggest that these pairs are related: (LAN, P600), (LAN, EPNP), (LAN, PNP), (ELAN, N400), (ELAN, EPNP), (ELAN, PNP), (ELAN, P600), (EPNP, P600).}
  \label{fig:erp_results}
\end{figure*}

\begin{figure*}
  \centering
  \includegraphics[width=0.9\textwidth]{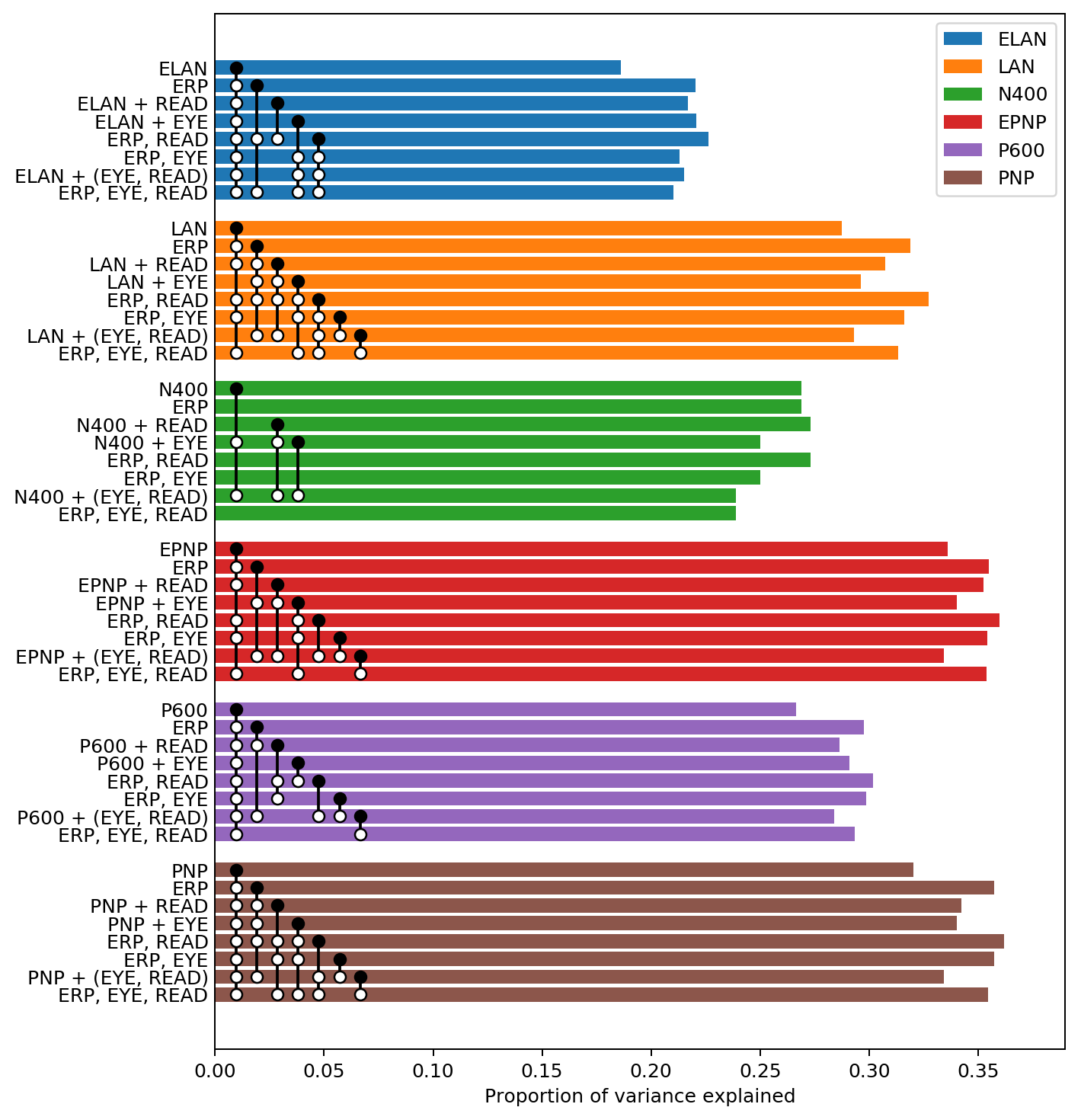}
  \caption[ERP results when behavioral data is included in training.]{The proportion of variance explained for prediction of each of the ERP signals (mean of 100 training runs). The target ERP is indicated by color; each group of bars shows performance for a different target ERP. The top bar in each group shows the proportion of variance explained when the model is trained using only the target ERP. Moving down, the next bar in each group, labeled \textit{ERP} shows the proportion of variance explained by the best combination of ERP signals for the target ERP. The other bars in each group moving from top to bottom show training variations that use behavioral data with either just the target ERP, or with the best combination of ERP signals. \textit{READ} denotes self-paced reading data, and \textit{EYE} denotes all four eye-tracking measures (in this analysis we use right-bounded pass time, gaze duration, go-past time, and first-fixation duration). Pairs of bars are significantly different from each other (paired t-test, false discovery rate < 0.01) if a black dot on one bar is connected to a white dot on the other bar by a contiguous vertical line. Self-paced reading time benefits prediction of all target ERP components except the N400. In the case of the ELAN, LAN, and PNP, self-paced reading time also has marginal benefit compared to the best combination of ERP training signals. Eye-tracking data benefits prediction of the ELAN, P600, and PNP components.}
  \label{fig:erp_with_eye}
\end{figure*}

Here we present a visualization (Figure \ref{fig:erp_results}) 
of the results presented in 
Table \ref{table:erp_table} 
of the main paper, and a visualization 
(Figure \ref{fig:erp_with_eye}) 
of a more complete set of results from which the information in 
Table \ref{table:behav_table} 
of the main paper is drawn. We also show supplemental results for variants of our primary analysis on multitask learning with eye-tracking, self-paced reading time and ERP data. In the variants we modify the input representation to our decoder network to see whether the relationships between the behavioral data and neural activity appear to be consistent with different choices of encoder architectures. Additional (and more varied) choices or architectures are left to future work. The results in Table \ref{table:behav_table_uni} reflect using only the forward-encoder (rather than the bi-LSTM) in the encoder network, while the results in Table \ref{table:behav_table_emb} reflect using only the word embeddings (i.e. bypassing the LSTM entirely). While the results are clearly worse for each of these choices of architecture than for using a bi-LSTM encoder, the relationships between the behavioral data and the ERP signals is qualitatively similar. Finally, \ref{table:raw_corr} shows the Pearson correlation coefficient between different measures. We note that the patterns of correlation are different than the patterns of which measures benefit from joint training with each other.

\begin{table*}
\centering
\begin{tabular}{l l c | l l c | l l c}
\toprule
Target & Additional & POVE & Target & Additional & POVE & Target & Additional & POVE \\
\midrule
ELAN & & 0.20 & LAN & & 0.23 & N400 & & 0.20 \\
ELAN & +ERP & \textbf{0.22} & LAN & + ERP & \textbf{0.26} & N400 & + ERP & 0.20 \\
ELAN & +READ & \textbf{0.22} & LAN & + READ & \textbf{0.25} & N400 & + READ & 0.20 \\
ELAN & +EYE & 0.21 & LAN & + EYE & \textbf{0.24} & N400 & + EYE & 0.18 \\
\midrule
EPNP & & 0.28 & P600 & & 0.24 & PNP & & 0.28 \\
EPNP & + ERP & 0.28 & P600 & + ERP & \textbf{0.25} & PNP & + ERP & \textbf{0.31} \\
EPNP & + READ & \textbf{0.29} & P600 & + READ & \textbf{0.25} & PNP & + READ & \textbf{0.30} \\
EPNP & + EYE & 0.29 & P600 & + EYE & 0.24 & PNP & + EYE & 0.29 \\
\bottomrule
\end{tabular}
\caption{Proportion of variance explained for each of the ERP components when using only the forward direction of the encoder (mean of 100 training runs). +ERP indicates the best combination of ERP training signals for the target ERP component, + READ indicates the inclusion of self-paced reading times, +EYE indicates the inclusion of eye-tracking data, and bold font indicates a significant difference from training on the target component alone.}
\label{table:behav_table_uni}
\end{table*}

\begin{table*}
\centering
\begin{tabular}{l l c | l l c | l l c}
\toprule
Target & Additional & POVE & Target & Additional & POVE & Target & Additional & POVE \\
\midrule
ELAN & & 0.15 & LAN & & 0.17 & N400 & & 0.05 \\
ELAN & + ERP & \textbf{0.18} & LAN & + ERP & \textbf{0.19} & N400 & + ERP & 0.05 \\
ELAN & + READ & \textbf{0.18} & LAN & + READ & \textbf{0.19} & N400 & + READ & 0.08 \\
ELAN & + EYE & \textbf{0.19} & LAN & + EYE & \textbf{0.19} & N400 & + EYE & 0.10 \\
\midrule
EPNP & & 0.18 & P600 & & 0.10 & PNP & & 0.20 \\
EPNP & + ERP & \textbf{0.20} & P600 & + ERP & \textbf{0.13} & PNP & + ERP & \textbf{0.23} \\
EPNP & + READ & \textbf{0.20} & P600 & + READ & \textbf{0.13} & PNP & + READ & \textbf{0.22} \\
EPNP & + EYE & \textbf{0.21} & P600 & + EYE & \textbf{0.14} & PNP & + EYE & \textbf{0.23} \\
\bottomrule
\end{tabular}
\caption{Proportion of variance explained for each of the ERP components when using only the word embeddings as input to the decoder and bypassing the LSTM entirely (mean of 100 training runs). +ERP indicates the best combination of ERP training signals for the target ERP component, + READ indicates the inclusion of self-paced reading times, +EYE indicates the inclusion of eye-tracking data, and bold font indicates a significant difference from training on the target component alone.}
\label{table:behav_table_emb}
\end{table*}

\begin{table*}
\centering
\begin{tabular}{l | c c c c c c c c c c c}
\toprule
Signal & ELAN & EPNP & LAN & N400 & P600 & PNP & FIX & PASS & GO & RIGHT & READ \\
\midrule
ELAN & 1.00 & 0.27 & 0.32 & 0.11 & 0.10 & 0.24 & 0.27 & 0.26 & 0.27 & 0.26 & -0.04 \\
EPNP & 0.27 & 1.00 & 0.66 & 0.41 & 0.50 & 0.83 & 0.17 & 0.17 & 0.19 & 0.17 & 0.02 \\
LAN & 0.32 & 0.66 & 1.00 & 0.58 & 0.33 & 0.47 & 0.12 & 0.11 & 0.13 & 0.12 & 0.01 \\
N400 & 0.11 & 0.41 & 0.58 & 1.00 & 0.47 & 0.33 & -0.04 & -0.04 & -0.02 & -0.03 & 0.11 \\
P600 & 0.10 & 0.50 & 0.33 & 0.47 & 1.00 & 0.69 & 0.14 & 0.14 & 0.16 & 0.14 & 0.10 \\
PNP & 0.24 & 0.83 & 0.47 & 0.33 & 0.69 & 1.00 & 0.25 & 0.24 & 0.26 & 0.25 & 0.03 \\
FIX & 0.27 & 0.17 & 0.12 & -0.04 & 0.14 & 0.25 & 1.00 & 1.00 & 1.00 & 1.00 & 0.04 \\
PASS & 0.26 & 0.17 & 0.11 & -0.04 & 0.14 & 0.24 & 1.00 & 1.00 & 1.00 & 1.00 & 0.04 \\
GO & 0.27 & 0.19 & 0.13 & -0.02 & 0.16 & 0.26 & 1.00 & 1.00 & 1.00 & 1.00 & 0.05 \\
RIGHT & 0.26 & 0.17 & 0.12 & -0.03 & 0.14 & 0.25 & 1.00 & 1.00 & 1.00 & 1.00 & 0.04 \\
READ & -0.04 & 0.02 & 0.01 & 0.11 & 0.10 & 0.03 & 0.04 & 0.04 & 0.05 & 0.04 & 1.00 \\
\bottomrule
\end{tabular}
\caption{Raw Pearson's correlation coefficients (computed on content words after the standardization and participant-averaging) between each neural and behavioral measure and each other measure. FIX indicates first-fixation time, PASS indicates first-pass time, GO indicates go-past time, RIGHT indicates right-bounded reading time, and READ indicates self-paced reading. Many of the measures are highly correlated, but the pattern of correlations is different from the pattern of benefits that we find during joint-training. In particular we note that the N400 is correlated with the other ERP signals, and yet we do not see benefit in prediction of the N400 when jointly training a model to predict it and other signals.}
\label{table:raw_corr}
\end{table*}

\end{document}